\documentclass[10pt,twocolumn,letterpaper]{article}

\usepackage{cvpr}              

\usepackage[dvipsnames]{xcolor}

\usepackage{graphicx}
\usepackage{booktabs}
\usepackage{multirow}
\usepackage{indentfirst}
\usepackage{makecell}

\definecolor{cvprblue}{rgb}{0.21,0.49,0.74}
\usepackage[pagebackref,breaklinks,colorlinks,citecolor=cvprblue]{hyperref}

\def\paperID{113} 
\def\confName{3DV\xspace}
\def\confYear{2025\xspace}

\title{RGBD-Glue: General Feature Combination for Robust RGB-D Point Cloud Registration}

\author{
Congjia Chen$^{1}$ \qquad
Xiaoyu Jia$^{2}$ \qquad
Yanhong Zheng$^{2}$ \qquad
Yufu Qu$^{1}$ \\
$^{1}$Beihang University \qquad
$^{2}$Beijing Institute of Spacecraft System Engineering
}

\begin{document}
\maketitle
\begin{abstract}
Point cloud registration is a fundamental task for estimating rigid transformations between point clouds. Previous studies have used geometric information for extracting features, matching and estimating transformation. Recently, owing to the advancement of RGB-D sensors, researchers have attempted to combine visual and geometric information to improve registration performance. However, these studies focused on extracting distinctive features by deep feature fusion, which cannot effectively solve the negative effects of each feature's weakness, and cannot sufficiently leverage the valid information. In this paper, we propose a new feature combination framework, which applies a looser but more effective combination. An explicit filter based on transformation consistency is designed for the combination framework, which can overcome each feature's weakness. And an adaptive threshold determined by the error distribution is proposed to extract more valid information from the two types of features. Owing to the distinctive design, our proposed framework can estimate more accurate correspondences and is applicable to both hand-crafted and learning-based feature descriptors. Experiments on ScanNet and 3DMatch show that our method achieves a state-of-the-art performance.
\end{abstract}    
\section{Introduction}
\label{sec:intro}


Point cloud registration, which aligns partial views of a scene, is a key component in numerous tasks such as SLAM and robotics applications. Researchers typically use methods that rely on correspondence extraction and geometric fitting to estimate the geometric transformation between two point clouds. Here, correspondence extraction is crucial, as high-quality correspondences can effectively improve the accuracy of geometric fitting. 
\begin{figure}[tp]
    \centering
    \includegraphics[width=\linewidth]{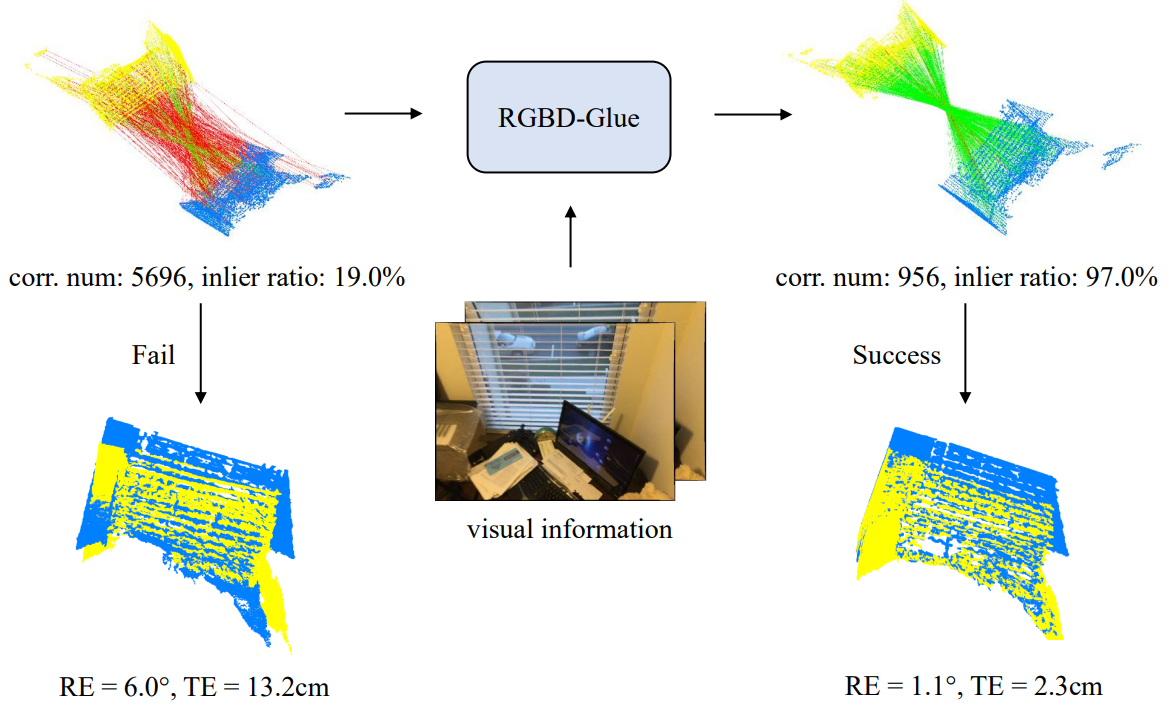}
    \caption{RGBD-Glue combines visual and geometric features to estimate credible correspondences for geometric fitting, which can achieve low rotation errors (REs) and translation errors (TEs) in registration. }
    \label{fig:sketch}
\end{figure}
\par
Correspondence extraction can be categorized into two components: feature extraction and correspondence estimation. For feature extraction, a representative hand-crafted feature descriptor is FPFH\cite{fpfh}, and recent learning-based feature descriptors\cite{spinnet, d3feat, fcgf, predator} have been improved significantly by learning more distinctive features. However, even with the learning-based feature descriptors, the low inlier ratio is still a hard problem to any correspondence-based geometric fitting methods. To this end, correspondence estimation methods are used to find more inliers. Recent studies have used CNN\cite{dgr}, PointNet\cite{learningmultiview} or attention module\cite{vbreg} to improve the inlier ratio. Although these methods are effective, they incur additional training costs, and their generalization is not always promising.
\par
The aforementioned methods are based only on geometric information. Owing to the rapid popularization of inexpensive RGB-D cameras, the acquisition of RGB-D data has become easier, thus facilitating the use of visual information. In recent studies\cite{unsupervisedrr, byoc, llt, pointmbf}, visual information has been used for point cloud registration, which demonstrated a state-of-the-art performance. However, UR\&R\cite{unsupervisedrr} uses only point clouds for localization and does not fully leverage geometric information. Meanwhile, BYOC\cite{byoc} uses visual correspondences as labels to bootstrap feature learning, which does not sufficiently exploit the correlation between visual and geometric features. LLT\cite{llt} and PointMBF\cite{pointmbf} extract fused visual-geometric features to deeply combine the two modalities, but this tight strategy will allow the weakness of one feature to affect another, and cannot sufficiently leverage the valid information of visual and geometric features. Besides, they all focus on training a feature extraction model using both two modalities, which have limited their generalization and flexibility. 
\par
In this paper, we aim to propose a method that can combine the two modalities in a more effective and flexible way. Visual feature descriptors can generate more distinctive keypoints and features than geometric feature descriptors, and can achieve a higher accuracy in feature matching\cite{byoc}. For this reason, visual correspondences can easily achieve a roughly accurate transformation estimation. However, sparse matches and mapping errors limit their performance, making it hard to achieve fine registration. By contrast, geometric feature descriptors are more stable to generate dense correspondences, but heavy outliers pose a considerable challenge to geometric fitting. Inspired by these, we propose a general feature combination framework RGBD-Glue, our key insight is to utilize the complementary properties of visual and geometric features, combine their advantages to overcome each feature's weakness.
\par
As mentioned before, it's easier to estimate a roughly accurate transformation via visual correspondences, which can be served as an useful prior information. Using the transformation, we estimate the error distribution of the assumed inliers and design an adaptive threshold for distribution test. An explicit filter is then proposed to extract the geometric correspondences with high transformation consistency and obtain a credible correspondence set. Finally, accurate registration can be achieved via the set. Compared with previous RGB-D combination studies\cite{byoc, unsupervisedrr, llt, pointmbf}, our proposed method can effectively combine the advantages of visual and geometric features, and avoids performance degradation when one of the two features is weak. Besides, our method focuses on flexibility and makes feature combination in a looser but more effective way, which brings our method unique advantages and outstanding performance. As shown in \cref{fig:sketch}, the correct correspondences are obtained and an accurate registration is achieved.
\par
Owing to the design on feature combination, our proposed framework is compatible with any visual and geometric feature descriptors, and can achieve better performance compared with using individual feature. Benefit from this, our proposed framework can use diverse feature descriptors for various tasks rather than depending on particular networks, and can be learning-free when needed.
\par
For evaluation, we conduct experiments on RGB-D datasets ScanNet\cite{scannet} and 3DMatch\cite{3dmatch}, and compare the performance of our method with those of recent point cloud registration methods. Besides, we conduct comprehensive ablation studies to demonstrate the effects of each component in our method.
\par
In summary, the contributions of this study are as follows:
\par
$\bullet$ We propose a flexible framework that combines visual and geometric features in a simple and novel way to achieve better point cloud registration. The experiments show great improvement than individual geometric or visual feature.
\par
$\bullet$ We propose a loose feature combination strategy and design an explicit filter based on transformation consistency for reliable correspondence estimation.
\section{Related Work}
\label{sec:related}

\subsection{Image Feature Matching}
\par
Image feature matching uses local features that are sparse keypoints associated with a descriptor to obtain the correspondences between two images. Classical methods, which rely on hand-crafted keypoint detectors and feature descriptors\cite{sift, surf, orb, brisk}, have been proven to be effective. Meanwhile, learning-based methods used in recent studies\cite{lift, superpoint, d2net} have demonstrated outstanding performances.
\par
Local features are typically matched with a nearest-neighbor search in the feature space. In this process, non-matchable keypoints and imperfect features usually result in mismatches. To solve this issue, Lowe’s ratio test\cite{sift} and RANSAC\cite{ransac} are commonly used to identify inliers. Recently, the excellent performance of learning-based methods has been demonstrated\cite{loftr, aspanformer, matchformer, superglue, lightglue}. These methods commonly use self-attention and cross-attention to aggregate information for more accurate matching, can obtain reliable correspondences when sufficiently trained. Also, the recent studies have demonstrated remarkable generalization of the learning-based methods\cite{loftr, lightglue}, making the pretrained networks transferable for unseen data. In this study, we leverage the high quality of visual features, utilize visual correspondences to improve the performance of RGB-D point cloud registration.

\subsection{Point Cloud Registration}
\par
Classical methods are ICP-based\cite{icp, gicp, nicp}. They consider the nearest neighbor points as correspondences and estimate the geometric transformation iteratively using an optimization algorithm. In recent studies, most methods have focused on feature extraction\cite{fpfh, fcgf, d3feat, predator} and correspondence estimation\cite{mac, cofinet, geotransformer, sc2pcr, vbreg}, where transformation is solved using a robust algorithm\cite{ransac, teaser}.
\par
FPFH\cite{fpfh} is one of the best hand-crafted descriptors. And several learning-based descriptors\cite{fcgf, d3feat, predator, cofinet} with advanced performance have been proposed. Most recently, GeoTransformer\cite{geotransformer} apply geometric attention aggregation to capture global contexts and encode geometric structure, which demonstrates remarkable performance. However, these descriptors are based only on geometric information. Meanwhile, the popularization of RGB-D cameras renders the utilization of visual information promising. To the best of our knowledge, BYOC\cite{byoc} demonstrates the manner by which visual correspondences can improve geometric feature learning, and UR\&R\cite{unsupervisedrr} uses RGB-D data to realize unsupervised learning. Subsequently, LLT\cite{llt} and PointMBF\cite{pointmbf} propose fusion networks to extract fused visual-geometric features. Our study is inspired by these methods. However, instead of extracting more distinctive features, we focus on achieving better correspondence estimation using the extracted features.
\par
The main purpose of correspondence estimation is to obtain accurate correspondences for geometric fitting. Early methods ranked correspondences based on feature similarity scores\cite{mian2005automatic} or voting scores\cite{glent2014search, sahloul2020accurate, yang2023mutual}. Chen et al.\cite{sc2pcr} proposed SC2-PCR, which uses a second-order spatial compatibility metric to rank correspondences. Zhang et al.\cite{mac} proposed MAC, which uses maximal cliques to obtain consistent sets and generate accurate pose hypotheses. In addition, PointDSC\cite{pointdsc} and VBReg\cite{vbreg} train networks to remove outliers. Compared with these methods, our method can achieve better performance by leveraging both visual and geometric information.

\subsection{RGB-D Combination}
\par
RGB-D data have two modalities: depth image, which can provide geometric information, and RGB image, which contains visual information. Several methods have been proposed to leverage both modalities. In typical point cloud tasks such as detection\cite{deepfusion, pointpainting}, segmentation\cite{bidirectional, fusenet}, and registration\cite{unsupervisedrr, byoc, llt, pointmbf}, the combination of visual and geometric information improves the performance.
\par
For registration, UR\&R\cite{unsupervisedrr} directly uses visual information instead of geometric information for feature extraction and matching. BYOC\cite{byoc} uses visual information to create pseudo-correspondence labels that replace pose supervision and achieve self-supervised registration. LLT\cite{llt} uses a local linear transformation module to embed geometric features into visual features and extract fused visual-geometric features. PointMBF\cite{pointmbf} combines visual and geometric information with bidirectional fusion, so that visual and geometric feature information can interact with each others, and extracts deep fused features. However, these methods use visual correspondences to create labels for better feature learning, or extract fused visual-geometric feature to deeply combine the two modalities, which cannot effectively combine the advantages of each feature. Differ from these methods, our proposed method uses a looser combination strategy, which can effectively exploit the complementary advantages of visual and geometric features.
\section{Method}
\label{sec:method}

\par
\begin{figure*}[tp]
    \centering
    \includegraphics[width=\linewidth]{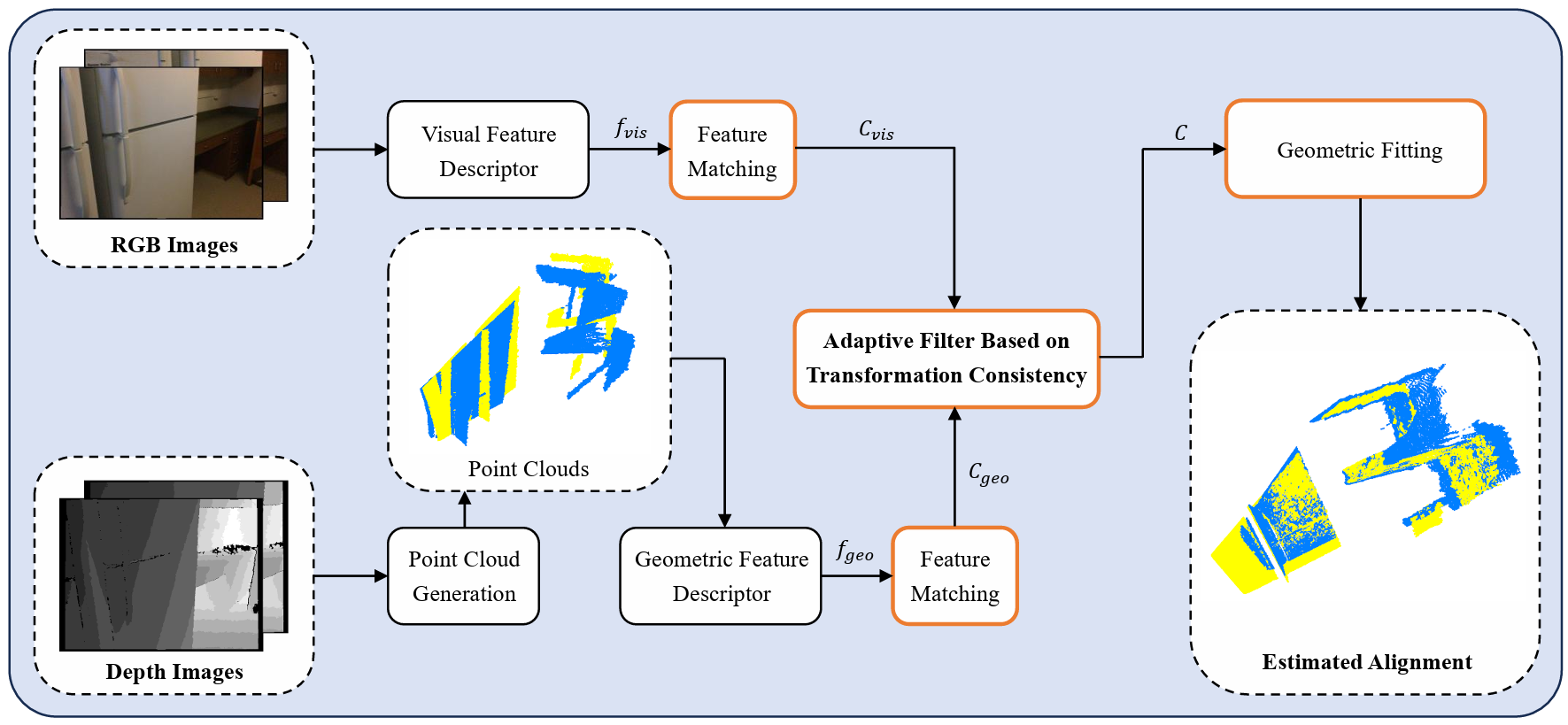}
    \caption{Architecture of the proposed RGBD-Glue framework. First, we extract both visual and geometric features from RGB-D data. Second, we match them to obtain correspondences. Third, we leverage the high-quality visual correspondences to find credible geometric correspondences by testing the transformation consistency based on an adaptive threshold. Finally, we estimate the transformation via the correspondences. }
    \label{fig:framework}
\end{figure*}
The framework is illustrated in \cref{fig:framework}. We first extract visual and geometric features respectively and match the features to get correspondence sets $C_{vis}$ and $C_{geo}$. Then, we propose an adaptive filter based on transformation consistency to extract valid information and obtain a credible set $C$. Finally, we use the set $C$ to estimate the transformation using same geometric fitting method with BYOC\cite{byoc} and relevant RGB-D works\cite{unsupervisedrr, llt, pointmbf}. The details of each component are presented in the following sections.

\subsection{Feature Matching}
\par
In our preprocessing step, we split the RGB-D image pair into RGB images $I_{0},I_{1} \in \mathbb{R}^{H \times W \times 3}$ and point clouds $P_{0},P_{1} \in \mathbb{R}^{N \times 3}$. Subsequently, we use an image descriptor and a point cloud descriptor to extract visual and geometric features $\mathbf{f}_{vis}, \mathbf{f}_{geo}$ from the data. After feature extraction, we match the features by directly finding the nearest neighbor to each point in the appropriate feature space. Consequently, we obtain a set of visual correspondences $C_{vis}$ and a set of geometric correspondences $C_{geo}$.
\par
To improve the inlier ratio, we filter the visual correspondences using Lowe’s ratio\cite{sift}. Given a visual correspondence pair $\mathbf{p}, \mathbf{q}$ and the corresponding feature vectors $\mathbf{f}_{p}, \mathbf{f}_{q}$, we calculate the ratio as follows:
\begin{equation} \label{eq1}
r_{p,q} = \frac{D(\mathbf{f}_{p},\mathbf{f}_{q})}{D(\mathbf{f}_{p},\mathbf{f}_{q}^{nn2})},
\end{equation}
where $D(\cdot)$ is the Euclidean distance in the feature space, $\mathbf{f}_{q}$ is the nearest neighbor of $\mathbf{f}_{p}$ and $\mathbf{f}_{q}^{nn2}$ is the second-nearest neighbor for $\mathbf{f}_{p}$. The correspondences whose ratio exceeds a predefined threshold will be regarded as unreliable points and thus eliminated from the subsequent steps. Note that, for matching methods such as LightGlue\cite{lightglue}, owing to their own feature matchers, we directly use the output correspondences without the ratio test.

\subsection{Adaptive Filter Based on Transformation Consistency} \label{sec:3.2}
In this process, we first estimate the geometric transformation via visual correspondences and obtain assumed inliers to estimate the error distribution. An adaptive threshold is then calculated through the estimated distribution. Using the threshold, we filter the geometric correspondences and extract the credible ones to obtain the set $C$, which contains the valid information and is used to achieve accurate geometric fitting. \\
\textbf{Estimation of the error distribution.} By mapping the visual correspondences into 3D points using the camera intrinsic matrix, we can turn the set $C_{vis}$ into a 3D correspondence set. Then, we use RANSAC to estimate a roughly accurate geometric transformation $\mathbf{T}$ between two point clouds via the set $C_{vis}$. Given the estimated transformation $\mathbf{T}$ and a matching pair $\mathbf{p}, \mathbf{q}$, we calculate the Euclidean distance error $d_{T}$ as follows:
\begin{equation} \label{eq2}
d_{T} = \Vert \mathbf{T}(\mathbf{p})-\mathbf{q} \Vert ,
\end{equation}
where $\mathbf{T} (\mathbf{p})$ implies applying geometric transformation $\mathbf{T}$ to 3D point $\mathbf{p}$. For inliers conform to the transformation $\mathbf{T}$, we assume that the errors in the $x$-axis, $y$-axis and $z$-axis satisfy independent identical Gaussian distributions $N(0, \sigma^{2})$ and have same variance\cite{teaser}. Therefore, the distribution of the Euclidean distance error $d_{T}$ can be regard as a Chi-square distribution:
\begin{equation} \label{eq3}
\begin{aligned}
d_{T}^{2} = d_{x}^{2} + d_{y}^{2} + d_{z}^{2} \ \ , \ \ \frac{d_{x}^{2} + d_{y}^{2} + d_{z}^{2}}{\sigma^{2}} \sim \chi^{2}(3).
\end{aligned}
\end{equation}
To estimate the variance $\sigma^{2}$, we obtain assumed inliers in $C_{vis}$ using the inlier threshold $t_{in}$, which has also been used in the RANSAC transforamtion estimation. An inlier set $C_{vis}^{in}$ can be obtained with the assumed inliers:
\begin{equation} \label{eq4}
C_{vis}^{in} = \{ \mathbf{p}_{i},\mathbf{q}_{i} \in C_{vis} | \Vert \mathbf{T}(\mathbf{p}_{i})-\mathbf{q}_{i} \Vert \leqslant K t_{in} \} .
\end{equation}
where the multiplier $K$ is introduced to tolerate the rough transformation estimation. When the visual correspondences are weak, the estimated transformation is likely to incorrect. In this case, the assumed inliers obtained by the RANSAC threshold $t_{in}$ can not reflect the real estimation error, and the estimated variance will be lower than the true value. For real-world data, making the inlier threshold of variance estimation larger than the RANSAC threshold can improve the robustness. After that, we use the set $C_{vis}^{in}$ as a sample for moment estimation:
\begin{equation} \label{eq5}
\hat{\sigma}^{2} = \frac{1}{3 \vert C_{vis}^{in} \vert} \sum_{i}^{ \vert C_{vis}^{in} \vert } \Vert \mathbf{T}(\mathbf{p}_{i}) - \mathbf{q}_{i} \Vert ^{2} .
\end{equation}
\textbf{Calculation of the adaptive threshold.} It's obvious that outliers do not conform to the geometric transformation, and thus have a considerably larger distance error than inliers. Therefore, if we have estimated the error distribution of inliers, we can regard the correspondences which have low probability of conforming to the distribution as outliers. To this end, we establish the 95\% confidence interval of the estimated distribution:
\begin{equation} \label{eq6}
\begin{aligned}
P( \frac{d^{2}}{\sigma^{2}} \leqslant \chi^{2}_{0.95}(3) ) = 0.95,
\end{aligned}
\end{equation}
where $\chi^{2}_{0.95}(3)$ is the 0.95 quantile of Chi-square distribution. The correspondences whose distance error is within the confidence interval have a high probability of conforming to the error distribution, and are likely to be the inliers. Therefore, we can calculate the adaptive threshold $\epsilon$:
\begin{equation} \label{eq7}
\epsilon = \sqrt{\sigma^{2} \chi^{2}_{0.95}(3)}.
\end{equation}
This threshold can be used to find the correspondences with high transformation consistency and inlier probability. Since the threshold is calculated from the error distribution, the value of threshold depends on the quality of visual correspondences. Weak visual correspondences usually result in rough transformation estimation and large error. In this case, a loose threshold will be calculated. On the contrary, strong visual correspondences usually have a minor error and result in a strict threshold. Therefore, the designed threshold is adaptive that it can vary suitably with different situations, which is important for the robustness.
\begin{figure}[tp]
    \centering
    \includegraphics[width=\linewidth]{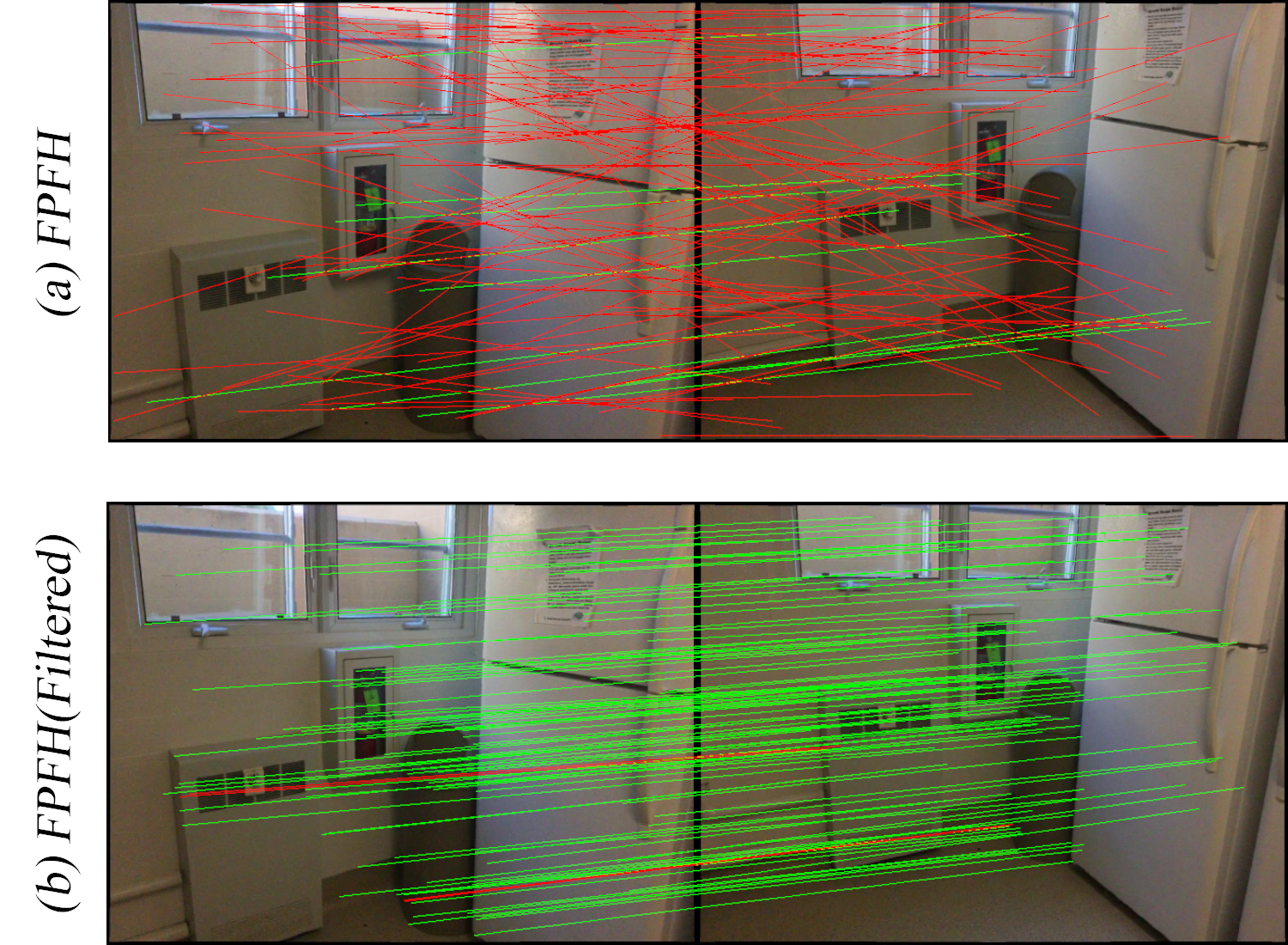}
    \caption{Correspondence estimation result yielded by our proposed method. The green lines denote the correct correspondences, and the red lines denote the incorrect correspondences. For better visualization, we randomly sample parts of matches to draw the lines. The result shows that the processed correspondences have high inlier ratio.}
    \label{fig:filter}
\end{figure} \\
\textbf{Extraction of credible geometric correspondences.} Using the threshold $\epsilon$, we can apply a test to determine the geometric correspondences that conform to the estimated distribution:
\begin{equation} \label{eq8}
C_{geo}^{in} = \{ \mathbf{p}_{i},\mathbf{q}_{i} \in C_{geo} | \Vert \mathbf{T}(\mathbf{p}_{i})-\mathbf{q}_{i} \Vert \leqslant \epsilon \}.
\end{equation}
After filtering, the survived geometric correspondences will have a distance error within the confidence interval. This indicates that the corresponding transformation of the survived correspondences is likely to be consistent with the estimated transformation $\mathbf{T}$, and thus those correspondences are more likely to be the correct matches. Therefore, we can effectively remove the incorrect matches by the filter. The filter results are shown in \cref{fig:filter}. This combination strategy aims to overcome the weakness of each type of feature. We leverage the high accuracy of visual correspondences to find credible geometric correspondences and remove the outliers. Meanwhile, the abundant information extracted from geometric correspondences can significantly improve the registration accuracy. Therefore, we can effectively leverage the advantages of both two types of features, and obtain sufficient and credible correspondences of two point clouds. To avoid the negative effect of wrong transformation estimation, the filter will be skipped when there are few visual correspondences or when there are almost no survived correspondences after filtering. Finally, we merge the visual correspondence set $C_{vis}^{in}$ and the geometric correspondence set $C_{geo}^{in}$ to obtain the final set $C$:
\begin{equation} \label{eq9}
C = C_{vis}^{in} + C_{geo}^{in}.
\end{equation}

\subsection{Geometric Fitting} \label{sec:3.3}
\par
Using the set $C$, we re-estimate the transformation $\mathbf{T} \in SE(3)$ that minimizes the error between the aligned correspondences in set $C$ as follows:
\begin{equation} \label{eq10}
\mathbf{T} = \mathop{argmin} \limits_{\mathbf{T}^{'} \in SE(3)} \frac{1}{\vert C \vert}  \sum_{(\mathbf{p},\mathbf{q},w) \in C} w_{i} \Vert \mathbf{T}^{'} (\mathbf{p}_{i}) - \mathbf{q}_{i} \Vert ^{2},
\end{equation}
where $w$ is a weight computed using the feature distance of $\mathbf{p},\mathbf{q}$. This equation can be solved as a weighted Procrustes problem\cite{generalizedprocrustes, kabsch1976solution, sorkine2017least}, which is commonly used in point cloud registration methods such as BYOC\cite{byoc}. Moreover, we adopt the same randomized optimization with BYOC and relevant studies\cite{unsupervisedrr, llt, pointmbf} to improve robustness.
\section{Experiment}
\label{sec:experiment}

\begin{table*}[tp]
\centering
\tabcolsep = 4.3pt
\fontsize{8}{10}\selectfont
\begin{tabular}{*{17}{c}}
    \toprule
    \multirow{3}{*}{} & \multirow{3}{*}{\makecell{Point Cloud \\ Train Set}} & \multicolumn{5}{c}{Rotation(deg)} & \multicolumn{5}{c}{Translation(cm)} & \multicolumn{5}{c}{Chamfer(mm)} \\
    & & \multicolumn{3}{c}{Accuracy $\uparrow$} & \multicolumn{2}{c}{Error $\downarrow$} & \multicolumn{3}{c}{Accuracy $\uparrow$} & \multicolumn{2}{c}{Error $\downarrow$} & \multicolumn{3}{c}{Accuracy $\uparrow$} & \multicolumn{2}{c}{Error $\downarrow$} \\
    \cmidrule(lr){3-5} \cmidrule(lr){6-7} \cmidrule(lr){8-10} \cmidrule(lr){11-12} \cmidrule(lr){13-15} \cmidrule(lr){16-17}
    & & 5 & 10 & 45 & Mean & Med. & 5 & 10 & 25 & Mean & Med. & 1 & 5 & 10 & Mean & Med. \\
    \midrule
    FCGF\cite{fcgf}+MAC\cite{mac} & \multirow{10}{*}{3DMatch} & 95.3 & 98.2 & \underline{99.1} & 3.1 & 1.3 & 67.3 & 88.5 & 97.1 & 7.4 & 3.5 & 82.4 & 95.1 & 97.1 & 5.9 & \underline{0.2} \\
    FCGF\cite{fcgf}+VBReg\cite{vbreg} & & \underline{96.8} & \underline{98.6} & \underline{99.1} & \underline{2.7} & 1.0 & 80.5 & \underline{94.0} & \underline{97.8} & \underline{6.0} & 2.6 & 90.2 & \underline{97.2} & \underline{97.9} & \underline{4.4} & \underline{0.2} \\
    GeoTransformer\cite{geotransformer} & & 94.0 & 96.8 & 98.1 & 4.3 & 1.0 & 79.2 & 92.0 & 96.7 & 8.2 & 2.5 & 88.4 & 95.8 & 96.9 & 5.8 & \textbf{0.1} \\
    PEAL\cite{peal} & & 94.4 & 96.8 & 98.4 & 3.9 & 0.9 & 80.5 & 92.8 & 97.0 & 7.3 & 2.4 & 89.1 & 96.0 & 97.1 & 6.0 & \textbf{0.1} \\
    BYOC\cite{byoc} & & 66.5 & 85.2 & 97.8 & 7.4 & 3.3 & 30.7 & 57.6 & 88.9 & 16.0 & 8.2 & 54.1 & 82.8 & 89.5 & 9.5 & 0.9 \\
    UR\&R\cite{unsupervisedrr} & & 87.6 & 93.1 & 98.3 & 4.3 & 1.0 & 69.2 & 84.0 & 93.8 & 9.5 & 2.8 & 79.7 & 91.3 & 94.0 & 7.2 & \underline{0.2} \\
    LLT\cite{llt} & & 93.4 & 96.5 & 98.8 & 3.0 & 0.9 & 76.9 & 90.2 & 96.7 & 6.4 & 2.4 & 86.4 & 95.1 & 96.8 & 5.3 & \textbf{0.1} \\
    PointMBF\cite{pointmbf} & & 94.6 & 97.0 & 98.7 & 3.0 & \underline{0.8} & \underline{81.0} & 92.0 & 97.1 & 6.2 & \underline{2.1} & \underline{91.3} & 96.6 & 97.4 & 4.9 & \textbf{0.1} \\
    Ours(SIFT+FCGF) & & 90.3 & 94.1 & 97.8 & 4.7 & 0.9 & 77.3 & 88.7 & 94.2 & 9.7 & 2.3 & 85.0 & 92.6 & 94.4 & 7.4 & \textbf{0.1} \\
    Ours(LightGlue+FCGF) & & \textbf{99.0} & \textbf{99.4} & \textbf{99.7} & \textbf{1.3} & \textbf{0.7} & \textbf{91.0} & \textbf{97.9} & \textbf{99.4} & \textbf{3.1} & \textbf{1.7} & \textbf{96.0} & \textbf{99.2} & \textbf{99.5} & \textbf{3.4} & \textbf{0.1} \\
    \midrule
    BYOC\cite{byoc} & \multirow{4}{*}{ScanNet} & 86.5 & 95.2 & 99.1 & 3.8 & 1.7 & 56.4 & 80.6 & 96.3 & 8.7 & 4.3 & 78.1 & 93.9 & 96.4 & 5.6 & 0.3 \\
    UR\&R\cite{unsupervisedrr} & & 92.7 & 95.8 & 98.5 & 3.4 & 0.8 & 77.2 & 89.6 & 96.1 & 7.3 & 2.3 & 86.0 & 94.6 & 96.1 & 5.9 & 0.1 \\
    LLT\cite{llt} & & 95.5 & 97.6 & 99.1 & 2.5 & 0.8 & 80.4 & 92.2 & 97.6 & 5.5 & 2.2 & 88.9 & 96.4 & 97.6 & 4.6 & 0.1 \\
    PointMBF\cite{pointmbf} & & 96.0 & 97.6 & 98.9 & 2.5 & 0.7 & 83.9 & 93.8 & 97.7 & 5.6 & 1.9 & 92.8 & 97.3 & 97.9 & 4.7 & 0.1 \\
    \bottomrule
\end{tabular}
\caption{Point cloud registration on ScanNet dataset. We use two visual features SIFT and LightGlue to evaluate our method, which are listed as Ours(SIFT+FCGF) and Ours(LightGlue+FCGF).}
\label{table1}
\end{table*}

\begin{table*}[tp]
\centering
\tabcolsep = 5.5pt
\fontsize{8}{10}\selectfont
\begin{tabular}{*{16}{c}}
    \toprule
    \multirow{3}{*}{} & \multicolumn{5}{c}{Rotation(deg)} & \multicolumn{5}{c}{Translation(cm)} & \multicolumn{5}{c}{Chamfer(mm)} \\
    & \multicolumn{3}{c}{Accuracy $\uparrow$} & \multicolumn{2}{c}{Error $\downarrow$} & \multicolumn{3}{c}{Accuracy $\uparrow$} & \multicolumn{2}{c}{Error $\downarrow$} & \multicolumn{3}{c}{Accuracy $\uparrow$} & \multicolumn{2}{c}{Error $\downarrow$} \\
    \cmidrule(lr){2-4} \cmidrule(lr){5-6} \cmidrule(lr){7-9} \cmidrule(lr){10-11} \cmidrule(lr){12-14} \cmidrule(lr){15-16}
    & 5 & 10 & 45 & Mean & Med. & 5 & 10 & 25 & Mean & Med. & 1 & 5 & 10 & Mean & Med. \\
    \midrule
    FCGF\cite{fcgf}+MAC\cite{mac} & 91.3 & 94.6 & 96.8 & 6.2 & 1.5 & 56.0 & 83.3 & 93.4 & 14.3 & 4.4 & 70.4 & 89.7 & 92.4 & 34.8 & \underline{0.4} \\
    FCGF\cite{fcgf}+VBReg\cite{vbreg} & \underline{94.1} & \underline{96.8} & \underline{97.3} & \underline{5.5} & \underline{1.2} & \underline{66.3} & \underline{87.2} & \underline{95.3} & \underline{11.3} & 3.6 & \underline{78.3} & \underline{93.1} & \underline{94.7} & 15.7 & \textbf{0.3} \\
    GeoTransformer\cite{geotransformer} & 92.0 & 95.3 & 96.9 & 6.1 & \underline{1.2} & 65.6 & 86.2 & 94.3 & 13.0 & 3.6 & 77.8 & 91.3 & 93.3 & \underline{12.4} & \textbf{0.3} \\
    PEAL\cite{peal} & 92.0 & 95.2 & 97.0 & 5.8 & \underline{1.2} & 66.1 & 86.8 & 94.6 & 12.3 & \underline{3.5} & 77.6 & 92.4 & 94.0 & 14.9 & \textbf{0.3} \\
    PointMBF\cite{pointmbf} & 86.9 & 91.7 & 96.3 & 6.9 & \underline{1.2} & 63.7 & 82.3 & 91.9 & 14.4 & \underline{3.5} & 75.0 & 89.1 & 91.5 & 13.9 & \textbf{0.3} \\
    Ours(SIFT+FCGF) & 86.2 & 89.8 & 94.1 & 10.1 & \underline{1.2} & 64.2 & 81.0 & 89.3 & 21.3 & 3.6 & 73.2 & 85.9 & 87.8 & 26.5 & \textbf{0.3} \\
    Ours(LightGlue+FCGF) & \textbf{96.4} & \textbf{98.2} & \textbf{99.0} & \textbf{2.6} & \textbf{1.0} & \textbf{72.9} & \textbf{91.5} & \textbf{96.8} & \textbf{6.7} & \textbf{2.9} & \textbf{83.0} & \textbf{94.5} & \textbf{96.4} & \textbf{6.5} & \textbf{0.3} \\
    \bottomrule
\end{tabular}
\caption{Point cloud registration on 3DMatch dataset. All learning-based point cloud methods are trained on 3DMatch dataset.}
\vspace{-0.1em}
\label{table2}
\end{table*}

\subsection{Experimental Settings}
\noindent
\textbf{Datasets.} We follow BYOC\cite{byoc} and use the large indoor RGB-D dataset ScanNet\cite{scannet} for evaluation. ScanNet is also used as the evaluation dataset for UR\&R\cite{unsupervisedrr}, LLT\cite{llt}, and PointMBF\cite{pointmbf}. Same to the case of BYOC, we use the official train/valid/test scene spilt and generate view pairs by sampling image pairs that are 20 frames apart. In addition, a smaller RGB-D dataset 3DMatch\cite{3dmatch} is used. As 3DMatch is a well-known dataset in point cloud registration tasks and has been used in previous studies\cite{mac, vbreg, geotransformer}, we use it as another evaluation dataset. \\
\textbf{Metrics.} We use the rotation error, the translation error and the chamfer distance to evaluate registration performance, report mean, median and accuracy under different thresholds, as in previous studies\cite{unsupervisedrr, byoc, llt, pointmbf}. We report the translation error in centimeters, the rotation error in degrees and the chamfer distance in millimeters. \\
\textbf{Baseline Methods.} We compare our method with recent state-of-the-art point cloud registration methods, including MAC\cite{mac}, VBReg\cite{vbreg}, GeoTransformer\cite{geotransformer}, and PEAL\cite{peal}. Since MAC and VBReg are correspondence estimation methods, we use FCGF\cite{fcgf} to extract putative correspondences for their estimation. The pretrained models used for VBReg, GeoTransformer, PEAL and FCGF are trained on 3DMatch dataset. Besides, we compare our method with recent feature combination methods, including UR\&R\cite{unsupervisedrr}, BYOC\cite{byoc}, LLT\cite{llt},  PointMBF\cite{pointmbf}. These methods are evaluated using models trained on 3DMatch dataset and ScanNet dataset respectively. \\
\textbf{Geometric Fitting.} We notice there are multiple geometric fitting methods used in the benchmark methods. For fairness, we use randomized WP mentioned in \cref{sec:3.3}, which is same with BYOC, UR\&R, LLT and PointMBF. For GeoTransformer and PEAL, we change their original fitting method to the same randomized WP using their output dense correspondences for obtaining fairer and more referable results. For MAC and VBReg, since their outputs are cliques or clusters which are not suitable for randomized WP, we maintain their own pipelines. Besides, no post-refinement is applied for our method and all benchmark methods, as same in BYOC, UR\&R, LLT, and PointMBF. \\
\textbf{Implementation Details.} As RGBD-Glue is a general framework that is compatible with any feature descriptor, we evaluate our method using different descriptors, including SIFT\cite{sift}, LightGlue\cite{lightglue}, FPFH\cite{fpfh}, and FCGF\cite{fcgf}. SIFT and FPFH are hand-crafted descriptors, FCGF is a learning-based descriptor, and LightGlue is a learning-based visual feature extractor and matcher. Experiments with more visual features can be found in supplementary material. For LightGlue, we directly use the pretrained model without fine-tuning, which is trained on image datasets Oxford \& Paris (for synthetic homography) and MegaDepth, and thus both ScanNet and 3DMatch datasets are unseen for LightGlue. Moreover, we downsample the point clouds with a voxel size of 2.5 cm, which is suitable for indoor datasets\cite{fcgf, pointcontrast, geotransformer}.

\begin{table*}[tp]
\centering
\tabcolsep = 7.0pt
\fontsize{8}{10}\selectfont
\begin{tabular}{*{16}{c}}
    \toprule
    \multirow{3}{*}{\makecell{Visual \\ Feature}} & \multicolumn{5}{c}{Rotation(deg)} & \multicolumn{5}{c}{Translation(cm)} & \multicolumn{5}{c}{Chamfer(mm)} \\
    & \multicolumn{3}{c}{Accuracy $\uparrow$} & \multicolumn{2}{c}{Error $\downarrow$} & \multicolumn{3}{c}{Accuracy $\uparrow$} & \multicolumn{2}{c}{Error $\downarrow$} & \multicolumn{3}{c}{Accuracy $\uparrow$} & \multicolumn{2}{c}{Error $\downarrow$} \\
    \cmidrule(lr){2-4} \cmidrule(lr){5-6} \cmidrule(lr){7-9} \cmidrule(lr){10-11} \cmidrule(lr){12-14} \cmidrule(lr){15-16}
    & 5 & 10 & 45 & Mean & Med. & 5 & 10 & 25 & Mean & Med. & 1 & 5 & 10 & Mean & Med. \\
    \midrule
    \multicolumn{16}{l}{\textbf{Only Visual Feature}} \\
    SIFT & 76.1 & 85.8 & 97.2 & 7.7 & 2.0 & 45.5 & 69.4 & 88.2 & 15.3 & 5.6 & 62.0 & 81.7 & 86.7 & 12.1 & 0.5 \\
    LightGlue & 98.2 & 99.4 & 99.8 & 1.7 & 1.2 & 72.0 & 95.7 & 99.4 & 4.4 & 3.4 & 90.6 & 98.9 & 99.3 & 3.8 & 0.2 \\
    \midrule
    \multicolumn{16}{l}{\textbf{FPFH + Visual Feature}} \\
    - &	24.2 & 56.4 & 92.8 & 18.3 & 8.7 & 5.5 & 19.3 & 55.7 & 39.9 & 22.0 & 8.8 & 35.5 & 53.6 & 27.9 & 8.7 \\
    SIFT & 86.9 & 91.4 & 96.9 & 6.4 & 0.9 & 73.6 & 84.8 & 92.0 & 12.4 & 2.4 & 80.9 & 89.5 & 91.7 & 9.7 & 0.1 \\
    LightGlue & 98.8 & 99.4 & 99.6 & 1.4 & 0.7 & 92.3 & 98.1 & 99.3 & 3.1 & 1.6 & 96.3 & 99.1 & 99.3 & 3.3 & 0.1 \\
    \midrule
    \multicolumn{16}{l}{\textbf{FCGF + Visual Feature}} \\
    - & 53.6 & 77.3 & 95.4 & 12.0 & 4.5 & 23.8 & 43.5 & 73.9 & 27.5 & 12.2 & 32.3 & 62.0 & 74.4 & 18.5 & 2.7 \\
    SIFT & 90.3 & 94.1 & 97.8 & 4.7 & 0.9 & 77.3 & 88.7 & 94.2 & 9.7 & 2.3 & 85.0 & 92.6 & 94.4 & 7.4 & 0.1 \\
    LightGlue & 99.0 & 99.4 & 99.7 & 1.3 & 0.7 & 91.0 & 97.9 & 99.4 & 3.1 & 1.7 & 96.0 & 99.2 & 99.5 & 3.4 & 0.1 \\
    \bottomrule
\end{tabular}
\caption{Point cloud registration with different settings. We evaluate the transformation estimated by only visual correspondences without subsequent process using geometric correspondences. And we remove the filter and evaluate the transformation estimated by the original geometric correspondences. Results with different combinations of features are also provided.}
\label{table3}
\end{table*}

\subsection{Registration on ScanNet Dataset}
\par
Following BYOC and PointMBF, we use 3DMatch dataset as the point cloud train set for comparison with geometry-only methods. As shown in \cref{table1}, our proposed method achieves a state-of-the-art performance in all the metrics. And when uses typical SIFT as the visual descriptor, our method also shows competitive performance. By effectively leveraging the advantages of both visual and geometric features, our method achieves the best performance among correspondence estimation methods and other feature combination methods. In particular, our method has a remarkable accuracy under the strict thresholds, and an accuracy of approximately 100\% under the loose thresholds. Notably, our method is a general framework compatible with any feature descriptor. Therefore, our method is more flexible for handling various RGB-D point cloud registration tasks.

\subsection{Registration on 3DMatch Dataset}
\par
3DMatch and 3DLoMatch used in previous geometry-only methods are processed from original 3DMatch dataset by merging multi-frame point clouds\cite{geotransformer, mac, vbreg}, which is not suitable for RGB-D methods like PointMBF due to their image-based process. Therefore, we sample image pairs same as in ScanNet for evaluation. We use the same train/valid/test scene spilt as in previous studies. Besides, we notice 20 frames apart is too easy, making the difference between methods unobvious. Therefore, we generate view pairs with lower overlap by sampling image pairs that are 40 frames apart, which can make the registration more challenging. We campare our method with MAC\cite{mac}, VBReg\cite{vbreg}, GeoTransformer\cite{geotransformer}, PEAL\cite{peal}, and the representative deep-fused network PointMBF\cite{pointmbf}. All methods are trained on 3DMatch dataset. As shown in \cref{table2}, our method still achieves the best performance.

\subsection{Ablations}
\par
We conduct comprehensive ablation studies on ScanNet dataset to demonstrate the effectiveness of our proposed method. To make further demonstration, we provide more experiment results in the supplementary material.
\begin{table}[tp]
\centering
\tabcolsep = 1.5pt
\fontsize{8}{12}\selectfont
\begin{tabular}{*{9}{c}}
    \toprule
    \multirow{2}{*}{\makecell{Geometric \\ Feature}} & \multirow{2}{*}{\makecell{Visual \\ Feature}} & \multicolumn{3}{c}{Mean Inlier Ratio(\%)} & \multicolumn{3}{c}{Mean Inlier Amount} & \multirow{2}{*}{\makecell{Filter \\ Recall}} \\
    \cmidrule(lr){3-5} \cmidrule(lr){6-8}
    & & 10 cm & 5 cm & 2.5 cm & 10 cm & 5 cm & 2.5 cm & \\
    \midrule
    \multirow{3}{*}{FPFH} & - & 11.5 & 7.2 & 3.8 & 693 & 424 & 217 & - \\
    & SIFT & 76.1 & 48.6 & 23.5 & 599 & 404 & 210 & 96.0 \\
    & LightGlue & 96.9 & 80.9 & 45.1 & 380 & 340 & 207 & 99.4 \\
    \midrule
    \multirow{3}{*}{FCGF} & - & 50.1 & 34.9 & 15.2 & 3502 & 2383 & 1013 & - \\
    & SIFT & 85.8 & 60.0 & 25.5 & 3127 & 2254 & 973 & 96.0 \\
    & LightGlue & 97.5 & 82.3 & 41.5 & 2303 & 1973 & 968 & 99.4 \\
    \bottomrule
\end{tabular}
\caption{Inlier ratio and amount of the filtered geometric correspondences. We filter the original geometric correspondences using our method and find inliers under different thresholds. 10 cm is a typical inlier threshold\cite{pointdsc, geotransformer}. Filter recall denotes the percentage of data processed by filter.}
\vspace{-1em}
\label{table4}
\end{table} \\
\textbf{Visual features vs. geometric features.} We observe that visual features can achieve better registration performance, LightGlue with no fine-tuning can already achieve great feature matching. Similar case has also been reported in BYOC\cite{byoc}. This may be because of the abundant texture information of visual features, which renders them more distinctive than geometric features. However, this does not imply that visual features are better in all cases. Because of varying viewpoints, different lighting, and weak texture, generating sufficient inliers using only visual features is unstable. By contrast, geometric features have different challenges, such as ambiguous and repetitive structures, and the low inlier ratio is usually a more serious problem. As shown in \cref{table3} and \cref{table4}, although most geometric correspondences are outliers, there are still quite a number of inliers that can achieve accurate registration after filtering.
\begin{table}[tp]
\centering\
\tabcolsep = 3.5pt
\fontsize{8}{12}\selectfont
\begin{tabular}{*{11}{c}}
    \toprule
    \multirow{2}{*}{$K$} & \multicolumn{3}{c}{Rotation(deg)} & \multicolumn{3}{c}{Translation(cm)} & \multicolumn{3}{c}{Inlier Ratio(\%)} & \multirow{2}{*}{\makecell{Filter \\ Recall}} \\
    \cmidrule(lr){2-4} \cmidrule(lr){5-7} \cmidrule(lr){8-10}
    & 5 & 10 & 45 & 5 & 10 & 25 & 10 & 5 & 2.5 &  \\
    \midrule
    \multicolumn{11}{l}{\textbf{SIFT + FCGF}} \\
    1 & 84.1 & 91.1 & 97.6 & 66.5 & 80.0 & 90.7 & 86.8 & 77.1 & 46.2 & 81.9 \\
    3 & 88.5 & 93.2 & 97.9 & 75.7 & 86.6 & 93.3 & 88.3 & 68.8 & 31.9 & 92.0 \\
    5 & 90.3 & 94.1 & 97.8 & 77.3 & 88.7 & 94.2 & 85.7 & 60.4 & 26.3 & 95.9 \\
    7 & 90.6 & 94.2 & 97.6 & 76.8 & 89.2 & 94.4 & 81.3 & 55.2 & 23.9 & 97.2 \\
    \midrule
    \multicolumn{11}{l}{\textbf{LightGlue + FCGF}} \\
    1 & 98.8 & 99.3 & 99.6 & 90.2 & 97.6 & 99.2 & 98.4 & 91.3 & 56.4 & 99.2 \\
    3 & 99.0 & 99.4 & 99.7 & 91.0 & 97.9 & 99.4 & 97.5 & 83.1 & 44.8 & 99.5 \\
    5 & 99.0 & 99.5 & 99.7 & 90.9 & 97.9 & 99.5 & 96.4 & 79.2 & 40.7 & 99.5 \\
    7 & 98.9 & 99.5 & 99.7 & 90.3 & 97.9 & 99.5 & 95.5 & 76.2 & 38.1 & 99.5 \\
    \bottomrule
\end{tabular}
\caption{Registration accuracy, mean inlier ratio and filter recall under different $K$ values.}
\vspace{-1em}
\label{table5}
\end{table} 
\begin{table}[tp]
\centering\
\tabcolsep = 3pt
\fontsize{8}{12}\selectfont
\begin{tabular}{*{9}{c}}
    \toprule
    \multirow{3}{*}{} & \multicolumn{4}{c}{Rotation(deg)} & \multicolumn{4}{c}{Translation(cm)} \\
    & \multicolumn{3}{c}{Accuracy $\uparrow$} & \multicolumn{1}{c}{Error $\downarrow$} & \multicolumn{3}{c}{Accuracy $\uparrow$} & \multicolumn{1}{c}{Error $\downarrow$} \\
    \cmidrule(lr){2-4} \cmidrule(lr){5-5} \cmidrule(lr){6-8} \cmidrule(lr){9-9}
    & 5 & 10 & 45 & Med. & 5 & 10 & 25 & Med. \\
    \midrule
    \multicolumn{9}{l}{\textbf{40 frames apart}} \\
    PointMBF & 71.1 & 79.1 & 89.8 & 1.8 & 48.4 & 65.7 & 79.8 & 5.3 \\
    PEAL & 80.0 & 85.2 & 90.6 & 1.5 & 57.3 & 75.9 & 85.3 & 4.1 \\
    FCGF+VBReg & 87.0 & 91.1 & 93.5 & 1.5 & 58.5 & 79.7 & 89.0 & 4.1 \\
    Ours & \textbf{93.6} & \textbf{95.6} & \textbf{96.7} & \textbf{1.0} & \textbf{73.3} & \textbf{89.4} & \textbf{95.2} & \textbf{2.8} \\
    \midrule
    \multicolumn{9}{l}{\textbf{60 frames apart}} \\
    PointMBF & 45.4 & 55.8 & 73.3 & 6.6 & 25.9 & 38.4 & 55.1 & 18.8 \\
    PEAL & 62.5 & 68.6 & 77.2 & 2.5 & 39.1 & 57.2 & 67.8 & 7.2 \\
    FCGF+VBReg & 72.7 & 79.4 & 83.0 & 2.2 & 41.7 & 62.4 & 75.3 & 6.4 \\
    Ours & \textbf{81.4} & \textbf{85.1} & \textbf{88.3} & \textbf{1.5} & \textbf{55.0} & \textbf{74.1} & \textbf{84.0} & \textbf{4.4} \\
    \bottomrule
\end{tabular}
\caption{Registration accuracy and median error under large frame spacing. We use 40 and 60 frames apart to evaluate the performance on ScanNet dataset. And FCGF and LightGlue are used as the features of our method. }
\label{table6}
\vspace{-1em}
\end{table} \\
\textbf{Performance of feature combination.} As shown in \cref{table3}, our method shows a significant improvement compared with using individual geometric features. We observe that using visual features particularly LightGlue, can achieve a high registration performance. However, when combined with FPFH or FCGF, a much better performance can be achieved. More detailed, as shown in \cref{table4}, our method can allow the filtered geometric correspondences to achieve high inlier ratio and maintain most of the inliers under the 2.5 cm threshold. With high-quality visual correspondences, our method can fully extract the valid information from geometric correspondences. On the contrary, when visual correspondences are weak, our method can avoid incorrect filtering and retain most valid information, which prevents a worse performance. It's notable that LightGlue + FPFH gets similar registration performance with LightGlue + FCGF, while SIFT + FPFH is worse than SIFT + FCGF. It may be because the accurate prior transformation estimated by LightGlue has eliminated most effect of outliers. As shown in \cref{table4}, LightGlue + FPFH get even higher inlier ratio under 2.5 cm threshold than LightGlue + FCGF. Compared with other combination strategies\cite{llt, pointmbf}, our proposed strategy can effectively leverage the advantages of two different modalities and alleviate their respective limitations. Better prior transformation estimated by visual correspondences or more accurate geometric correspondences can both boost our method's performance. Moreover, as mentioned in \cref{sec:3.2}, filter will be skipped in some cases. Therefore, we additionally calculate the percentage of data that has been processed by the filter as filter recall. The results demonstrate that our proposed filter has effectively removed outliers in most of the data. \\
\textbf{Effect of \textit{\textbf{K}}.} As mentioned in \cref{sec:3.2}, we use a multiplier $K$ to magnify the threshold. This is because the visual correspondences may result in inaccurate transformation estimation when they are weak. Therefore, we set $K$ to reduce the constraints of visual feature for better robustness. The results are shown in \cref{table5}, which indicates that increasing $K$ results in better performance. More detailed, increasing $K$ can improve the filter's robustness when facing low-quality visual correspondences, but will also reduce the inlier ratio and cause slight performance deterioration when the visual correspondences are strong. We notice that the accuracy is insensitive to the variation of $K$ when $K$ is large. Therefore, we believe a fixed value set respectively for hand-crafted and learning-based visual features can handle general cases, and higher value may be needed when the visual correspondences are quite weak. In our all experiments, we set $K = 3$ for learning-based visual feature such as LightGlue, and $K = 5$ for hand-crafted visual feature such as SIFT. \\
\textbf{Registration under large frame spacing.} To further demonstrate our method's effectiveness and generalization, we evaluate the registration performance under larger frame spacing on ScanNet dataset. Larger frame spacing will cause lower overlap, which brings great challenge to both visual and geometric feature matching. We compare our method with VBReg\cite{vbreg}, PEAL\cite{peal} and PointMBF\cite{pointmbf}. The results are shown in \cref{table6}, which indicates that our method has significantly better performance.
\section{Conclusion}
\label{sec:conclusion}
\par
A general framework RGBD-Glue for RGB-D point cloud registration is proposed in this paper. Differ from other combination works, we use an explicit filter based on transformation consistency to achieve a looser combination and design an adaptive threshold to extract more valid information, which bring our method better performance and robustness. Besides, the applicability of different descriptors renders our proposed method flexible to various registration tasks. The experiments show that our method achieves a better performance compared with using individual feature, and demonstrates a state-of-the-art performance. Furthermore, exploiting more advantages from RGB images to improve correspondence estimation is still a worth exploring task. We will make further research in our future work.

{
    \small
    \bibliographystyle{ieeenat_fullname}
    \bibliography{main}

\begin{thebibliography}{49}
\providecommand{\natexlab}[1]{#1}
\providecommand{\url}[1]{\texttt{#1}}
\expandafter\ifx\csname urlstyle\endcsname\relax
  \providecommand{\doi}[1]{doi: #1}\else
  \providecommand{\doi}{doi: \begingroup \urlstyle{rm}\Url}\fi

\bibitem[Ao et~al.(2021)Ao, Hu, Yang, Markham, and Guo]{spinnet}
Sheng Ao, Qingyong Hu, Bo Yang, Andrew Markham, and Yulan Guo.
\newblock Spinnet: Learning a general surface descriptor for 3d point cloud registration.
\newblock In \emph{Proceedings of the IEEE/CVF conference on computer vision and pattern recognition}, pages 11753--11762, 2021.

\bibitem[Bai et~al.(2020)Bai, Luo, Zhou, Fu, Quan, and Tai]{d3feat}
Xuyang Bai, Zixin Luo, Lei Zhou, Hongbo Fu, Long Quan, and Chiew-Lan Tai.
\newblock D3feat: Joint learning of dense detection and description of 3d local features.
\newblock In \emph{Proceedings of the IEEE/CVF conference on computer vision and pattern recognition}, pages 6359--6367, 2020.

\bibitem[Bai et~al.(2021)Bai, Luo, Zhou, Chen, Li, Hu, Fu, and Tai]{pointdsc}
Xuyang Bai, Zixin Luo, Lei Zhou, Hongkai Chen, Lei Li, Zeyu Hu, Hongbo Fu, and Chiew-Lan Tai.
\newblock Pointdsc: Robust point cloud registration using deep spatial consistency.
\newblock In \emph{Proceedings of the IEEE/CVF Conference on Computer Vision and Pattern Recognition}, pages 15859--15869, 2021.

\bibitem[Bay et~al.(2006)Bay, Tuytelaars, and Van~Gool]{surf}
Herbert Bay, Tinne Tuytelaars, and Luc Van~Gool.
\newblock Surf: Speeded up robust features.
\newblock In \emph{Computer Vision--ECCV 2006: 9th European Conference on Computer Vision, Graz, Austria, May 7-13, 2006. Proceedings, Part I 9}, pages 404--417. Springer, 2006.

\bibitem[Besl and McKay(1992)]{icp}
Paul~J Besl and Neil~D McKay.
\newblock Method for registration of 3-d shapes.
\newblock In \emph{Sensor fusion IV: control paradigms and data structures}, pages 586--606. Spie, 1992.

\bibitem[Chen et~al.(2022{\natexlab{a}})Chen, Luo, Zhou, Tian, Zhen, Fang, Mckinnon, Tsin, and Quan]{aspanformer}
Hongkai Chen, Zixin Luo, Lei Zhou, Yurun Tian, Mingmin Zhen, Tian Fang, David Mckinnon, Yanghai Tsin, and Long Quan.
\newblock Aspanformer: Detector-free image matching with adaptive span transformer.
\newblock In \emph{European Conference on Computer Vision}, pages 20--36. Springer, 2022{\natexlab{a}}.

\bibitem[Chen et~al.(2022{\natexlab{b}})Chen, Sun, Yang, and Tao]{sc2pcr}
Zhi Chen, Kun Sun, Fan Yang, and Wenbing Tao.
\newblock Sc2-pcr: A second order spatial compatibility for efficient and robust point cloud registration.
\newblock In \emph{Proceedings of the IEEE/CVF Conference on Computer Vision and Pattern Recognition}, pages 13221--13231, 2022{\natexlab{b}}.

\bibitem[Choy et~al.(2019)Choy, Park, and Koltun]{fcgf}
Christopher Choy, Jaesik Park, and Vladlen Koltun.
\newblock Fully convolutional geometric features.
\newblock In \emph{Proceedings of the IEEE/CVF international conference on computer vision}, pages 8958--8966, 2019.

\bibitem[Choy et~al.(2020)Choy, Dong, and Koltun]{dgr}
Christopher Choy, Wei Dong, and Vladlen Koltun.
\newblock Deep global registration.
\newblock In \emph{Proceedings of the IEEE/CVF conference on computer vision and pattern recognition}, pages 2514--2523, 2020.

\bibitem[Dai et~al.(2017)Dai, Chang, Savva, Halber, Funkhouser, and Nie{\ss}ner]{scannet}
Angela Dai, Angel~X Chang, Manolis Savva, Maciej Halber, Thomas Funkhouser, and Matthias Nie{\ss}ner.
\newblock Scannet: Richly-annotated 3d reconstructions of indoor scenes.
\newblock In \emph{Proceedings of the IEEE conference on computer vision and pattern recognition}, pages 5828--5839, 2017.

\bibitem[DeTone et~al.(2018)DeTone, Malisiewicz, and Rabinovich]{superpoint}
Daniel DeTone, Tomasz Malisiewicz, and Andrew Rabinovich.
\newblock Superpoint: Self-supervised interest point detection and description.
\newblock In \emph{Proceedings of the IEEE conference on computer vision and pattern recognition workshops}, pages 224--236, 2018.

\bibitem[Dusmanu et~al.(2019)Dusmanu, Rocco, Pajdla, Pollefeys, Sivic, Torii, and Sattler]{d2net}
Mihai Dusmanu, Ignacio Rocco, Tomas Pajdla, Marc Pollefeys, Josef Sivic, Akihiko Torii, and Torsten Sattler.
\newblock D2-net: A trainable cnn for joint description and detection of local features.
\newblock In \emph{Proceedings of the ieee/cvf conference on computer vision and pattern recognition}, pages 8092--8101, 2019.

\bibitem[El~Banani and Johnson(2021)]{byoc}
Mohamed El~Banani and Justin Johnson.
\newblock Bootstrap your own correspondences.
\newblock In \emph{Proceedings of the IEEE/CVF International Conference on Computer Vision}, pages 6433--6442, 2021.

\bibitem[El~Banani et~al.(2021)El~Banani, Gao, and Johnson]{unsupervisedrr}
Mohamed El~Banani, Luya Gao, and Justin Johnson.
\newblock Unsupervisedr\&r: Unsupervised point cloud registration via differentiable rendering.
\newblock In \emph{Proceedings of the IEEE/CVF Conference on Computer Vision and Pattern Recognition}, pages 7129--7139, 2021.

\bibitem[Fischler and Bolles(1981)]{ransac}
Martin~A Fischler and Robert~C Bolles.
\newblock Random sample consensus: a paradigm for model fitting with applications to image analysis and automated cartography.
\newblock \emph{Communications of the ACM}, 24\penalty0 (6):\penalty0 381--395, 1981.

\bibitem[Glent~Buch et~al.(2014)Glent~Buch, Yang, Kruger, and Gordon~Petersen]{glent2014search}
Anders Glent~Buch, Yang Yang, Norbert Kruger, and Henrik Gordon~Petersen.
\newblock In search of inliers: 3d correspondence by local and global voting.
\newblock In \emph{Proceedings of the IEEE Conference on computer vision and pattern recognition}, pages 2067--2074, 2014.

\bibitem[Gojcic et~al.(2020)Gojcic, Zhou, Wegner, Guibas, and Birdal]{learningmultiview}
Zan Gojcic, Caifa Zhou, Jan~D Wegner, Leonidas~J Guibas, and Tolga Birdal.
\newblock Learning multiview 3d point cloud registration.
\newblock In \emph{Proceedings of the IEEE/CVF conference on computer vision and pattern recognition}, pages 1759--1769, 2020.

\bibitem[Gower(1975)]{generalizedprocrustes}
John~C Gower.
\newblock Generalized procrustes analysis.
\newblock \emph{Psychometrika}, 40:\penalty0 33--51, 1975.

\bibitem[Hazirbas et~al.(2017)Hazirbas, Ma, Domokos, and Cremers]{fusenet}
Caner Hazirbas, Lingni Ma, Csaba Domokos, and Daniel Cremers.
\newblock Fusenet: Incorporating depth into semantic segmentation via fusion-based cnn architecture.
\newblock In \emph{Computer Vision--ACCV 2016: 13th Asian Conference on Computer Vision, Taipei, Taiwan, November 20-24, 2016, Revised Selected Papers, Part I 13}, pages 213--228. Springer, 2017.

\bibitem[Hu et~al.(2021)Hu, Zhao, Jiang, Jia, and Wong]{bidirectional}
Wenbo Hu, Hengshuang Zhao, Li Jiang, Jiaya Jia, and Tien-Tsin Wong.
\newblock Bidirectional projection network for cross dimension scene understanding.
\newblock In \emph{Proceedings of the IEEE/CVF Conference on Computer Vision and Pattern Recognition}, pages 14373--14382, 2021.

\bibitem[Huang et~al.(2021)Huang, Gojcic, Usvyatsov, Wieser, and Schindler]{predator}
Shengyu Huang, Zan Gojcic, Mikhail Usvyatsov, Andreas Wieser, and Konrad Schindler.
\newblock Predator: Registration of 3d point clouds with low overlap.
\newblock In \emph{Proceedings of the IEEE/CVF Conference on computer vision and pattern recognition}, pages 4267--4276, 2021.

\bibitem[Jiang et~al.(2023)Jiang, Dang, Wei, Xie, Yang, and Salzmann]{vbreg}
Haobo Jiang, Zheng Dang, Zhen Wei, Jin Xie, Jian Yang, and Mathieu Salzmann.
\newblock Robust outlier rejection for 3d registration with variational bayes.
\newblock In \emph{Proceedings of the IEEE/CVF conference on computer vision and pattern recognition}, pages 1148--1157, 2023.

\bibitem[Kabsch(1976)]{kabsch1976solution}
Wolfgang Kabsch.
\newblock A solution for the best rotation to relate two sets of vectors.
\newblock \emph{Acta Crystallographica Section A: Crystal Physics, Diffraction, Theoretical and General Crystallography}, 32\penalty0 (5):\penalty0 922--923, 1976.

\bibitem[Leutenegger et~al.(2011)Leutenegger, Chli, and Siegwart]{brisk}
Stefan Leutenegger, Margarita Chli, and Roland~Y Siegwart.
\newblock Brisk: Binary robust invariant scalable keypoints.
\newblock In \emph{2011 International conference on computer vision}, pages 2548--2555. Ieee, 2011.

\bibitem[Li et~al.(2022)Li, Yu, Meng, Caine, Ngiam, Peng, Shen, Lu, Zhou, Le, et~al.]{deepfusion}
Yingwei Li, Adams~Wei Yu, Tianjian Meng, Ben Caine, Jiquan Ngiam, Daiyi Peng, Junyang Shen, Yifeng Lu, Denny Zhou, Quoc~V Le, et~al.
\newblock Deepfusion: Lidar-camera deep fusion for multi-modal 3d object detection.
\newblock In \emph{Proceedings of the IEEE/CVF Conference on Computer Vision and Pattern Recognition}, pages 17182--17191, 2022.

\bibitem[Lindenberger et~al.(2023)Lindenberger, Sarlin, and Pollefeys]{lightglue}
Philipp Lindenberger, Paul-Edouard Sarlin, and Marc Pollefeys.
\newblock Lightglue: Local feature matching at light speed.
\newblock In \emph{Proceedings of the IEEE/CVF International Conference on Computer Vision}, pages 17627--17638, 2023.

\bibitem[Lowe(2004)]{sift}
David~G Lowe.
\newblock Distinctive image features from scale-invariant keypoints.
\newblock \emph{International journal of computer vision}, 60:\penalty0 91--110, 2004.

\bibitem[Mian et~al.(2005)Mian, Bennamoun, and Owens]{mian2005automatic}
Ajmal~S Mian, Mohammed Bennamoun, and Robyn~A Owens.
\newblock Automatic correspondence for 3d modeling: An extensive review.
\newblock \emph{International Journal of Shape Modeling}, 11\penalty0 (02):\penalty0 253--291, 2005.

\bibitem[Qin et~al.(2023)Qin, Yu, Wang, Guo, Peng, Ilic, Hu, and Xu]{geotransformer}
Zheng Qin, Hao Yu, Changjian Wang, Yulan Guo, Yuxing Peng, Slobodan Ilic, Dewen Hu, and Kai Xu.
\newblock Geotransformer: Fast and robust point cloud registration with geometric transformer.
\newblock \emph{IEEE Transactions on Pattern Analysis and Machine Intelligence}, 2023.

\bibitem[Rublee et~al.(2011)Rublee, Rabaud, Konolige, and Bradski]{orb}
Ethan Rublee, Vincent Rabaud, Kurt Konolige, and Gary Bradski.
\newblock Orb: An efficient alternative to sift or surf.
\newblock In \emph{2011 International conference on computer vision}, pages 2564--2571. Ieee, 2011.

\bibitem[Rusu et~al.(2009)Rusu, Blodow, and Beetz]{fpfh}
Radu~Bogdan Rusu, Nico Blodow, and Michael Beetz.
\newblock Fast point feature histograms (fpfh) for 3d registration.
\newblock In \emph{2009 IEEE international conference on robotics and automation}, pages 3212--3217. IEEE, 2009.

\bibitem[Sahloul et~al.(2020)Sahloul, Shirafuji, and Ota]{sahloul2020accurate}
Hamdi Sahloul, Shouhei Shirafuji, and Jun Ota.
\newblock An accurate and efficient voting scheme for a maximally all-inlier 3d correspondence set.
\newblock \emph{IEEE Transactions on Pattern Analysis and Machine Intelligence}, 43\penalty0 (7):\penalty0 2287--2298, 2020.

\bibitem[Sarlin et~al.(2020)Sarlin, DeTone, Malisiewicz, and Rabinovich]{superglue}
Paul-Edouard Sarlin, Daniel DeTone, Tomasz Malisiewicz, and Andrew Rabinovich.
\newblock Superglue: Learning feature matching with graph neural networks.
\newblock In \emph{Proceedings of the IEEE/CVF conference on computer vision and pattern recognition}, pages 4938--4947, 2020.

\bibitem[Segal et~al.(2009)Segal, Haehnel, and Thrun]{gicp}
Aleksandr Segal, Dirk Haehnel, and Sebastian Thrun.
\newblock Generalized-icp.
\newblock In \emph{Robotics: science and systems}, page 435. Seattle, WA, 2009.

\bibitem[Serafin and Grisetti(2015)]{nicp}
Jacopo Serafin and Giorgio Grisetti.
\newblock Nicp: Dense normal based point cloud registration.
\newblock In \emph{2015 IEEE/RSJ International Conference on Intelligent Robots and Systems (IROS)}, pages 742--749. IEEE, 2015.

\bibitem[Sorkine-Hornung and Rabinovich(2017)]{sorkine2017least}
Olga Sorkine-Hornung and Michael Rabinovich.
\newblock Least-squares rigid motion using svd.
\newblock \emph{Computing}, 1\penalty0 (1):\penalty0 1--5, 2017.

\bibitem[Sun et~al.(2021)Sun, Shen, Wang, Bao, and Zhou]{loftr}
Jiaming Sun, Zehong Shen, Yuang Wang, Hujun Bao, and Xiaowei Zhou.
\newblock Loftr: Detector-free local feature matching with transformers.
\newblock In \emph{Proceedings of the IEEE/CVF conference on computer vision and pattern recognition}, pages 8922--8931, 2021.

\bibitem[Vora et~al.(2020)Vora, Lang, Helou, and Beijbom]{pointpainting}
Sourabh Vora, Alex~H Lang, Bassam Helou, and Oscar Beijbom.
\newblock Pointpainting: Sequential fusion for 3d object detection.
\newblock In \emph{Proceedings of the IEEE/CVF conference on computer vision and pattern recognition}, pages 4604--4612, 2020.

\bibitem[Wang et~al.(2022{\natexlab{a}})Wang, Zhang, Yang, Peng, and Stiefelhagen]{matchformer}
Qing Wang, Jiaming Zhang, Kailun Yang, Kunyu Peng, and Rainer Stiefelhagen.
\newblock Matchformer: Interleaving attention in transformers for feature matching.
\newblock In \emph{Proceedings of the Asian Conference on Computer Vision}, pages 2746--2762, 2022{\natexlab{a}}.

\bibitem[Wang et~al.(2022{\natexlab{b}})Wang, Huo, Chen, Zhang, Sheng, and Xu]{llt}
Ziming Wang, Xiaoliang Huo, Zhenghao Chen, Jing Zhang, Lu Sheng, and Dong Xu.
\newblock Improving rgb-d point cloud registration by learning multi-scale local linear transformation.
\newblock In \emph{European Conference on Computer Vision}, pages 175--191. Springer, 2022{\natexlab{b}}.

\bibitem[Xie et~al.(2020)Xie, Gu, Guo, Qi, Guibas, and Litany]{pointcontrast}
Saining Xie, Jiatao Gu, Demi Guo, Charles~R Qi, Leonidas Guibas, and Or Litany.
\newblock Pointcontrast: Unsupervised pre-training for 3d point cloud understanding.
\newblock In \emph{Computer Vision--ECCV 2020: 16th European Conference, Glasgow, UK, August 23--28, 2020, Proceedings, Part III 16}, pages 574--591. Springer, 2020.

\bibitem[Yang et~al.(2020)Yang, Shi, and Carlone]{teaser}
Heng Yang, Jingnan Shi, and Luca Carlone.
\newblock Teaser: Fast and certifiable point cloud registration.
\newblock \emph{IEEE Transactions on Robotics}, 37\penalty0 (2):\penalty0 314--333, 2020.

\bibitem[Yang et~al.(2023)Yang, Zhang, Fan, Ren, and Zhang]{yang2023mutual}
Jiaqi Yang, Xiyu Zhang, Shichao Fan, Chunlin Ren, and Yanning Zhang.
\newblock Mutual voting for ranking 3d correspondences.
\newblock \emph{IEEE Transactions on Pattern Analysis and Machine Intelligence}, 2023.

\bibitem[Yi et~al.(2016)Yi, Trulls, Lepetit, and Fua]{lift}
Kwang~Moo Yi, Eduard Trulls, Vincent Lepetit, and Pascal Fua.
\newblock Lift: Learned invariant feature transform.
\newblock In \emph{Computer Vision--ECCV 2016: 14th European Conference, Amsterdam, The Netherlands, October 11-14, 2016, Proceedings, Part VI 14}, pages 467--483. Springer, 2016.

\bibitem[Yu et~al.(2021)Yu, Li, Saleh, Busam, and Ilic]{cofinet}
Hao Yu, Fu Li, Mahdi Saleh, Benjamin Busam, and Slobodan Ilic.
\newblock Cofinet: Reliable coarse-to-fine correspondences for robust pointcloud registration.
\newblock \emph{Advances in Neural Information Processing Systems}, 34:\penalty0 23872--23884, 2021.

\bibitem[Yu et~al.(2023)Yu, Ren, Zhang, Zhou, Lin, and Dai]{peal}
Junle Yu, Luwei Ren, Yu Zhang, Wenhui Zhou, Lili Lin, and Guojun Dai.
\newblock Peal: Prior-embedded explicit attention learning for low-overlap point cloud registration.
\newblock In \emph{Proceedings of the IEEE/CVF Conference on Computer Vision and Pattern Recognition}, pages 17702--17711, 2023.

\bibitem[Yuan et~al.(2023)Yuan, Fu, Li, Meng, and Wang]{pointmbf}
Mingzhi Yuan, Kexue Fu, Zhihao Li, Yucong Meng, and Manning Wang.
\newblock Pointmbf: A multi-scale bidirectional fusion network for unsupervised rgb-d point cloud registration.
\newblock In \emph{Proceedings of the IEEE/CVF International Conference on Computer Vision}, pages 17694--17705, 2023.

\bibitem[Zeng et~al.(2017)Zeng, Song, Nie{\ss}ner, Fisher, Xiao, and Funkhouser]{3dmatch}
Andy Zeng, Shuran Song, Matthias Nie{\ss}ner, Matthew Fisher, Jianxiong Xiao, and Thomas Funkhouser.
\newblock 3dmatch: Learning local geometric descriptors from rgb-d reconstructions.
\newblock In \emph{Proceedings of the IEEE conference on computer vision and pattern recognition}, pages 1802--1811, 2017.

\bibitem[Zhang et~al.(2023)Zhang, Yang, Zhang, and Zhang]{mac}
Xiyu Zhang, Jiaqi Yang, Shikun Zhang, and Yanning Zhang.
\newblock 3d registration with maximal cliques.
\newblock In \emph{Proceedings of the IEEE/CVF Conference on Computer Vision and Pattern Recognition}, pages 17745--17754, 2023.

\end{thebibliography}
}

\clearpage
\setcounter{page}{1}
\maketitlesupplementary

In the supplementary material, we provide more experiment results to further demonstrate our method's effectiveness, and provide more detailed analysis about the performance. In \cref{sec:A}, we provide correspondence estimation results before and after the filtering under large frame spacing to further demonstrate our method's robustness. In \cref{sec:B}, we make feature combination with more visual features using our method to further demonstrate our method's flexibility and generalization. In \cref{sec:C}, we provide more analysis of our designed adaptive threshold to demonstrate its effectiveness.

\renewcommand{\thesection}{A}
\section{Filter's Performance Under Large Frame Spacing}
\label{sec:A}
\begin{figure}[tp]
    \centering
    \renewcommand{\thefigure}{S1}
    \includegraphics[width=\linewidth]{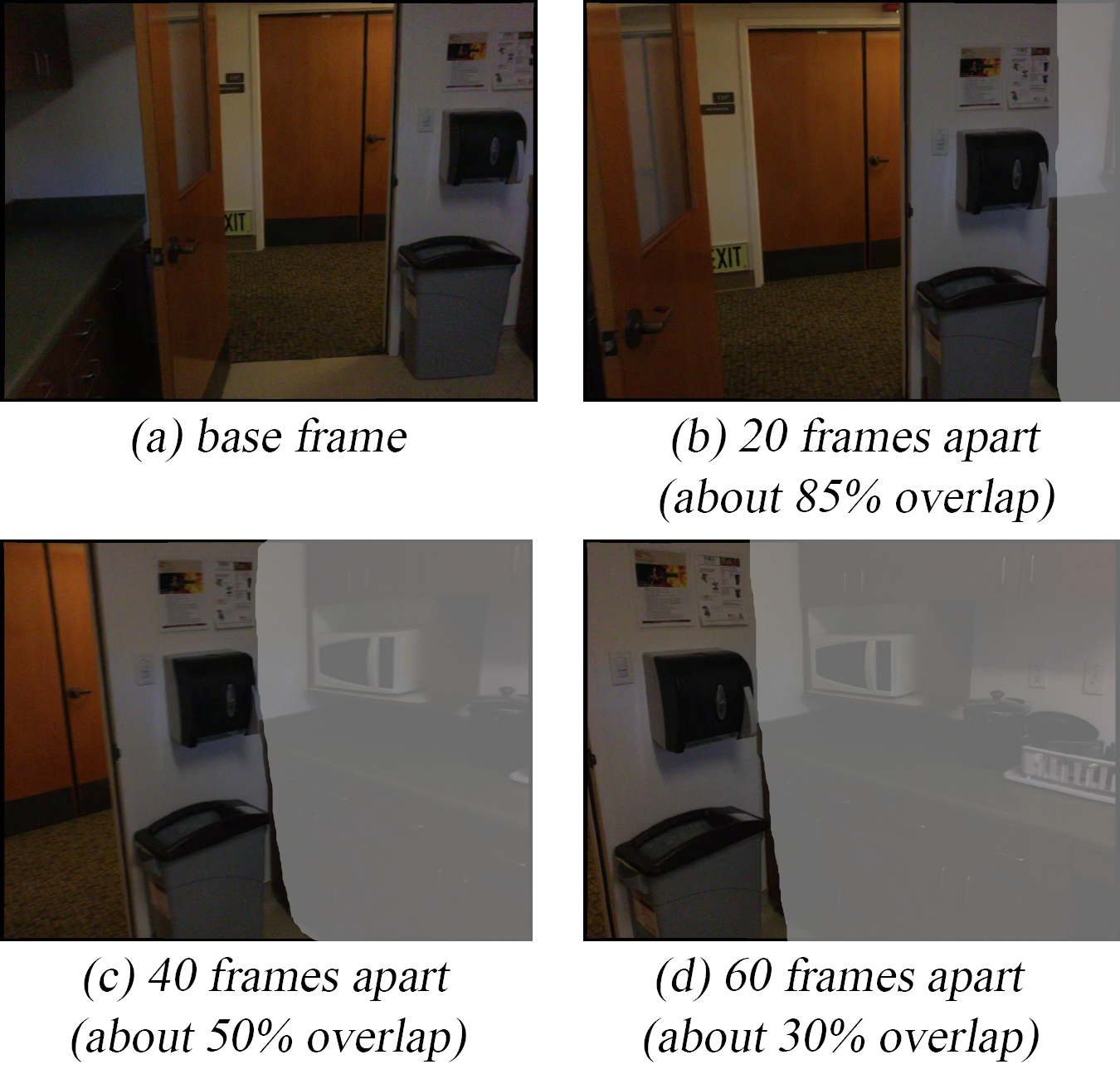}
    \caption{An example of different frame spacing. We reduce the opacity of non-overlap region for better visualization. Large frame spacing causes low overlap, which brings great challenge to both visual and geometric feature matching.}
    \label{fig:frame}
\end{figure}

As shown in \cref{fig:frame}, large frame spacing causes low overlap which deteriorates the performance of both visual and geometric feature matching. In this case, both visual and geometric correspondences get lower inlier ratio, which limits their registration performance. To make more detailed analysis about our method's performance under large frame spacing, we evaluate mean inlier ratio, mean inlier amount and filter recall after the filtering.
\par
The results are shown in \cref{tables1}, compare to the results under 20 frames apart in the main paper (\cref{table4}), the filtered correspondences under larger frame spacing have a less significant improvement on the inlier ratio. Except the effect of worse geometric correspondences and lower filter recall (data not processed by filter usually have low inlier ratios), sparser and worse visual correspondences can cause a larger error distribution and result in lower inlier ratio improvement. However, our method can still produce high-quality correspondences even with 60 frames apart, and thus outperform other methods in registration, as shown in the main paper (\cref{table6}). That demonstrates our method's strong robustness. Furthermore, filter recall is important to the filter's overall performance. It depends on the visual feature and the overlap of image pairs, a better or denser visual feature could improve the filter recall.
\begin{table}[tp]
\centering
\renewcommand{\thetable}{S1}
\tabcolsep = 3pt
\fontsize{8}{12}\selectfont
\begin{tabular}{*{8}{c}}
    \toprule
    \multirow{2}{*}{} & \multicolumn{3}{c}{Mean Inlier Ratio(\%)} & \multicolumn{3}{c}{Mean Inlier Amount} & \multirow{2}{*}{\makecell{Filter \\ Recall}} \\
    \cmidrule(lr){2-4} \cmidrule(lr){5-7}
    & 10 cm & 5 cm & 2.5 cm & 10 cm & 5 cm & 2.5 cm & \\
    \midrule
    \multicolumn{8}{l}{\textbf{40 frames apart}} \\
    Before Filtering & 32.5 & 20.6 & 8.2 & 2242 & 1411 & 547 & - \\
    After Filtering & 91.2 & 70.1 & 30.5 & 1563 & 1232 & 529 & 96.5 \\
    \midrule
    \multicolumn{8}{l}{\textbf{60 frames apart}} \\
    Before Filtering & 22.2 & 13.5 & 5.2 & 1521 & 918 & 340 & - \\
    After Filtering & 78.8 & 56.6 & 22.8 & 1085 & 813 & 329 & 88.4 \\
    \bottomrule
\end{tabular}
\caption{Mean inlier ratio and mean inlier amount under large frame spacing. FCGF and LightGlue are used without Add Corr. Filter recall denotes the percentage of data processed by filter.}
\vspace{-1em}
\label{tables1}
\end{table}


\begin{table*}[tp]
\centering
\renewcommand{\thetable}{S2}
\tabcolsep = 6pt
\fontsize{8}{10}\selectfont
\begin{tabular}{*{16}{c}}
    \toprule
    \multirow{3}{*}{\makecell{Visual \\ Feature}} & \multicolumn{5}{c}{Rotation(deg)} & \multicolumn{5}{c}{Translation(cm)} & \multicolumn{5}{c}{Chamfer(mm)} \\
    & \multicolumn{3}{c}{Accuracy $\uparrow$} & \multicolumn{2}{c}{Error $\downarrow$} & \multicolumn{3}{c}{Accuracy $\uparrow$} & \multicolumn{2}{c}{Error $\downarrow$} & \multicolumn{3}{c}{Accuracy $\uparrow$} & \multicolumn{2}{c}{Error $\downarrow$} \\
    \cmidrule(lr){2-4} \cmidrule(lr){5-6} \cmidrule(lr){7-9} \cmidrule(lr){10-11} \cmidrule(lr){12-14} \cmidrule(lr){15-16}
    & 5 & 10 & 45 & Mean & Med. & 5 & 10 & 25 & Mean & Med. & 1 & 5 & 10 & Mean & Med. \\
    \midrule
    \multicolumn{16}{l}{\textbf{FPFH + Visual Feature}} \\
    BRISK & 85.7 & 90.0 & 96.7 & 6.8 & 0.9 & 73.3 & 84.4 & 91.3 & 12.5 & 2.4 & 80.0 & 88.1 & 90.9 & 9.5 & 0.1 \\
    ORB & 81.8 & 86.7 & 94.3 & 10.1 & 1.0 & 68.8 & 80.1 & 88.2 & 17.9 & 2.6 & 76.0 & 84.6 & 87.3 & 13.5 & 0.2 \\
    SuperGlue & 98.7 & 99.2 & 99.6 & 1.5 & 0.7 & 91.3 & 97.8 & 99.3 & 3.4 & 1.7 & 95.8 & 99.0 & 99.3 & 3.3 & 0.1 \\
    \midrule
    \multicolumn{16}{l}{\textbf{FCGF + Visual Feature}} \\
    BRISK & 89.3 & 93.2 & 97.5 & 5.3 & 0.9 & 76.5 & 88.0 & 93.7 & 10.3 & 2.3 & 84.0 & 91.9 & 93.8 & 8.1 & 0.1 \\
    ORB & 87.2 & 91.8 & 96.3 & 6.9 & 0.9 & 74.5 & 85.6 & 92.0 & 12.8 & 2.4 & 81.6 & 89.7 & 91.8 & 9.9 & 0.1 \\
    SuperGlue & 98.8 & 99.4 & 99.7 & 1.4 & 0.7 & 90.2 & 97.9 & 99.4 & 3.2 & 1.8 & 95.8 & 99.1 & 99.4 & 3.4 & 0.1 \\
    \bottomrule
\end{tabular}
\caption{Point cloud registration with more visual features. BRISK\cite{brisk} and ORB\cite{orb} are hand-crafted features, SuperGlue\cite{superglue} is a learning-base feature. The evaluation is made on ScanNet dataset.}
\label{tables2}
\end{table*}

\renewcommand{\thesection}{B}
\section{Combination with More Visual Features}
\label{sec:B}
To further demonstrate our method's generalization and flexibility, we use BRISK\cite{brisk}, ORB\cite{orb} and SuperGlue\cite{superglue} to make extra experiments and evaluate the registration performance on ScanNet dataset. BRISK and ORB are hand-crafted features. SuperGlue is similar with LightGlue\cite{lightglue} since LightGlue is an improvement work of SuperGlue which mainly have improvement on efficiency, and we choose SuperPoint\cite{superpoint} as the feature descriptor for both two methods. We combine those visual features with FPFH or FCGF using our proposed method and evaluate their registration performance on ScanNet dataset. The results are shown in \cref{tables2}. 
\par
The results show that our framework is compatible with various visual features, and better visual feature descriptor or matcher can bring better performance. For practical application, it's easy to change the feature descriptor of our framework for the more suitable one. For example, use a faster hand-crafted descriptor for better efficiency, or use a deeper learning-based descriptor for better accuracy. Benefit from this, our method is more flexible and general than other combination methods in practical tasks.


\renewcommand{\thesection}{C}
\section{Variation of The Adaptive Threshold}
\label{sec:C}
To further demonstrate the effectiveness of our design, we make additional experiments to check the variation of the adaptive threshold. We generate view pairs by sampling image pairs that are 20 frames apart on ScanNet dataset, use LightGlue to extract visual correspondences and calculate the adaptive threshold. Since there are almost no data that have a squared threshold value exceed 0.01 when $K=3$, we divide the squared threshold value between 0 and 0.01 into five ranges. For each range, we find the view pairs that have a calculated threshold corresponding to the range, and get their mean visual match amount. Furthermore, we map the visual correspondences into 3D correspondences and check the inliers under different distance thresholds. The comparison is made with mean match amount and mean inlier ratio, since denser and more accurate visual correspondences result in a more accurate estimation of geometric transformation.
\par
As shown in \cref{tables3}, higher threshold means sparser visual correspondences and lower inlier ratio, which demonstrates that our designed adaptive threshold can change properly under different qualities of visual correspondences. When the visual correspondences are dense and accurate, the adaptive threshold will be strict to sufficiently remove the geometric outliers. On the contrary, when the visual correspondences are sparse and inaccurate, the estimated geometric transformation will get error, and the adaptive threshold will be loose to avoid removing the correct geometric correspondences. Benefit form this, our method can handle various situations without extra parameters, and can avoid the negative effect of weak visual feature. Using our designed adaptive threshold, the filter can fully extract valid information from the geometric correspondences when visual correspondences are strong, and can preserve most valid information when visual correspondences are weak, which avoids a worse registration performance than individual feature.
\begin{table}[tp]
\centering
\renewcommand{\thetable}{S3}
\tabcolsep = 8pt
\fontsize{8}{12}\selectfont
\begin{tabular}{*{5}{c}}
    \toprule
    \multirow{2}{*}{\makecell{Threshold Range \\ (squared value)}} & \multirow{2}{*}{\makecell{Mean Match \\ Amount}} & \multicolumn{3}{c}{Mean Inlier Ratio(\%)} \\
    \cmidrule(lr){3-5}
    & & 10 cm & 5 cm & 2.5 cm \\
    \midrule
    0 - 0.002 & 277 & 97.5 & 95.2 & 84.1 \\
    0.002 - 0.004 & 251 & 95.4 & 88.1 & 65.6 \\
    0.004 - 0.006 & 238 & 91.8 & 78.3 & 50.7 \\
    0.006 - 0.008 & 237 & 87.5 & 68.9 & 40.5 \\
    0.008 - 0.01 & 227 & 80.7 & 58.7 & 31.4 \\
    \bottomrule
\end{tabular}
\caption{Mean visual match amount and mean inlier ratio of different threshold ranges. LightGlue\cite{lightglue} is used as the visual feature.}
\vspace{-1em}
\label{tables3}
\end{table}

\end{document}